\documentclass[letterpaper, 10 pt, conference]{ieeeconf}  

\IEEEoverridecommandlockouts                              

\usepackage{graphics} 
\usepackage{epsfig} 
\usepackage{mathptmx} 
\usepackage{times} 
\usepackage{amsmath} 
\usepackage{amssymb}  
\usepackage{algorithm} 
\usepackage{cite}
\usepackage{algpseudocode} 
\usepackage{lipsum}
\usepackage{graphicx}
\usepackage[dvipsnames]{xcolor}
\usepackage{eucal}
\usepackage{mathtools,leftidx}
\usepackage{subcaption}
\usepackage{wrapfig}
\usepackage{bbm}
\usepackage{amsmath}

\usepackage{multirow}

\usepackage{color, colortbl}
\usepackage{xcolor}
\usepackage[bookmarks=true]{hyperref}
\usepackage{bm}

\usepackage[utf8]{inputenc}
\usepackage{fontenc}
\usepackage{siunitx}
\usepackage{tikz}
\usepackage{textcomp}
\usepackage{flushend}

\newcommand\footnoteref[1]{\protected@xdef\@thefnmark{\ref{#1}}\@footnotemark}

\vspace{-20pt}
\title{\LARGE \bf
 Visual-Geometry GP-based Navigable Space for Autonomous Navigation
 }
\vspace{-20pt}

\makeatletter
\newcommand*\titleheader[1]{\gdef\@titleheader{#1}}
\AtBeginDocument{%
  \let\st@red@title\@title
  \def\@title{%
    \bgroup\normalfont\large\centering\@titleheader\par\egroup
    \vskip1.5em\st@red@title}
}
\makeatother

\vspace{-20pt}

\author{\vspace{-10pt}Mahmoud Ali, Durgkant Pushp, Zheng Chen, and Lantao Liu
\vspace{-20pt}
\thanks{
All authors are with the Luddy School of Informatics, Computing, and Engineering, Indiana University, Bloomington, IN 47408 USA.  {\tt\small \{alimaa, dpushp, zc11, lantao\}@iu.edu}} 
}

\titleheader{  \textcolor{gray}{This paper has been accepted for publication at  2024 IEEE/RSJ International Conference on Intelligent Robots and Systems} \textcolor{blue}{\href{https://iros2024-abudhabi.org/}{( IROS 2024)}} }

\begin{document}

\newcommand\ZC[1]{\textcolor{blue}{Zheng: #1}}

\maketitle
\thispagestyle{empty}
\pagestyle{empty}

\begin{abstract}  
Autonomous navigation in unknown environments is challenging and demands the consideration of both geometric and semantic information in order to parse the navigability of the environment. 
In this work, we propose a novel space modeling framework, Visual-Geometry Sparse Gaussian Process (VG-SGP), that simultaneously considers  semantics and geometry of the scene. Our proposed approach can overcome the limitation of visual planners that fail to recognize geometry associated with the semantic and the geometric planners that completely overlook the semantic information which is very critical in real-world navigation. 
The proposed method leverages dual Sparse Gaussian Processes in an integrated manner; the first is trained to forecast geometrically navigable spaces while the second predicts the semantically navigable areas. This integrated model is able to pinpoint the overlapping (geometric and semantic) navigable space.
The simulation and real-world experiments demonstrate that the ability of the proposed VG-SGP model, coupled with our innovative navigation strategy, outperforms models solely reliant on visual or geometric navigation algorithms, highlighting a superior adaptive behavior. 
We provided a demonstration video\footnote{Video: \href{https://youtu.be/0s6VSj5Z1dg}{https://youtu.be/0s6VSj5Z1dg}} and open-sourced our code
\footnote{Code: \href{https://github.com/mahmoud-a-ali/vg-nav}{https://github.com/mahmoud-a-ali/vg-nav}}.
\end{abstract}
\vspace{-4pt}
\section{Introduction and Related Work}
\vspace{-4pt}
It is a challenging task for autonomous robots to navigate in real-world, unpredictable field environments which typically contain considerable complexity, with obstacles that may include large rocks, shrubs, and tree stumps, as well as terrains of mixed composition such as asphalt, sand, and mud.
When deploying a robot in such a setting, it must determine the navigable portions of the scene it captures. 
One approach to identifying navigable space involves utilizing the geometry or structure of surrounding objects, without considering their semantic interpretation. This geometric method typically leverages a 3D representation of the environment, such as point clouds or depth observations~\cite{weerakoon2022terp, liang2022adaptiveon}.
Geometric methodologies commonly employ global or local representation of the navigable space of the environment. 
Global representations include the use of an occupancy grid map, which encodes the likelihood of occupancy within a 2D spatial domain~\cite{thrun2003learning}, a 2.5D elevation map\cite{Fankhauser2016GridMapLibrary} that outlines the terrain elevation, or 3D maps that uses voxels to model the 3D space, such as Octomaps~\cite{hornung2013octomap}.
On the other hand, local representations such as cost and traversability maps are used to aid local planners in devising navigational strategies~\cite{shin2021model}. 
Another approach to identifying the navigable spaces is the semantic approach which performs semantic segmentation of the scene to predict its traversability using visual input~\cite{zhu2017target, yang2018visual, cheng2022masked}. 
Traversability may be assigned manually based on terrain types, for example attributing low traversability to mud areas and high traversability to grass areas~\cite{ra27, ra33,ra34}. In contrast, recent learning-based approaches~\cite{ra11, ra12,ra13, ra32, ra35} circumvent heuristic methods by leveraging data to learn traversability costs.
\begin{figure} \vspace{3pt}
    \subfloat[]{%
    \includegraphics[width=0.23\textwidth,height=1.in]{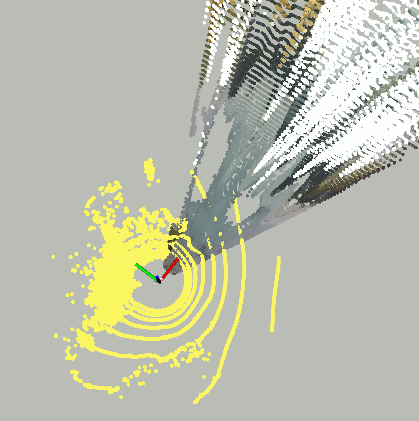}
    \label{fig_jnt_mdl_a}
   }  
    \subfloat[\label{fig_jnt_mdl_b}]{%
    \includegraphics[width=0.23\textwidth, height=1.in]{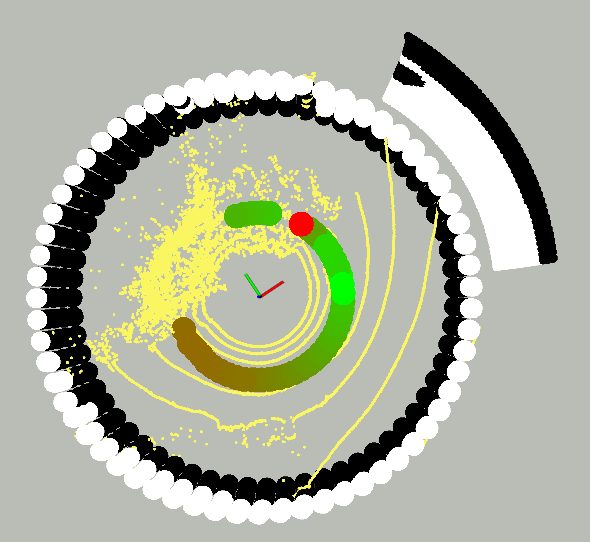}}
    \includegraphics[width=0.01\textwidth,height=1.in]{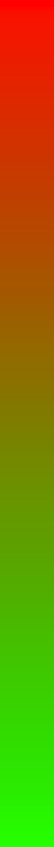}  \vspace{-8pt}
  \caption{\small 
  Visual-Geometry Navigable Space:  (a) shows LiDAR's pointcloud (yellow) and camera's colored-pointcloud; (b) shows the geometry navigable space (grey-coded circular surface, white regions represent free space), the visual navigable space (grey-coded vertical plane, white regions represents the navigable classes in the camera field of view), and the local navigation points  (colored-circles, where color represents the go-to-goal cost).
  \vspace{-5pt}} 
  \label{fig_jnt_mdl}
\end{figure}
Moreover, visual navigation has been approached through various methods like utilizing motion information from targets or optical flow features~\cite{OF_binoTrack, OF_ttt}, storing environment images with corresponding control actions~\cite{AB_Bista}, using image signal entropy for non-holonomic robot control~\cite{AB_Dame}, feature tracking~\cite{FT_Kim, mohamed2020model}, and path-finding through image segmentation based on control parameters from path boundaries~\cite{SB_Zhang}.
However, visual navigation may fall short in challenging environments where visually appealing navigable scenes (e.g. high-slope grassland) are non-navigable geometrically.
Gaussian Process (GP) has been used for modeling continuous spatial phenomena and creating navigability maps based on geometric~\cite{jadidi2017warped} or semantic interpretations~\cite{jadidi2017gaussian}.
While standard GPs are limited in real-time applications due to their high time complexity of $\mathcal{O}(n^3)$. Sparse GP (SGP) methods reduce GP's complexity to $\mathcal{O}(nm^2)$ by selecting $m$ inducing points to approximate the full dataset under Bayesian rules~\cite{snelson2006sparse,sheth2015sparse, titsias2009variational}.
LiDAR-based geometric navigation is more robust than visual-based approaches however, they may incorrectly identify navigable spaces due to a lack of semantic consideration.
While visual methods offer flexibility through pixel analysis but are limited by weather due to visual dependency. 
Recent navigation strategies merge geometric and visual data, employing end-to-end learning to analyze commands and trajectories across terrains~\cite{ra11, ra13}, but struggle with 3D terrain complexity due to the high cost of data analysis. Other methods use global maps to integrate semantic and geometric costs into a combined navigable space~\cite{ra, ra27}. 
In contrast, we propose a local mapless navigation approach that integrates visual and geometric navigable spaces using two local SGP models.

\section{Methodology} \label{methodology} \vspace{-4pt}
We propose a new framework that combines the geometry and the semantics of the robot's surroundings to identify the navigable spaces around the robot. 
Briefly, the output of the RGBD camera and LiDAR are processed in a parallel way where the RGB image is segmented by labeling each pixel with a unique class (grass, tree, asphalt, etc), and the pointcloud is represented by an occupancy surface around the robot. Consequentially, the segmented image is converted into a binary navigability image (navigable and non-navigable pixels) based on a defined set of navigable classes. The navigability is then projected on the camera's depth pointcloud which is represented as a \textit{Visual} SGP model (V-SGP) to find the (\textit{visual}) navigable spaces in front of the robot.
On the other hand, the occupancy surface is represented by a \textit{Geometry} SGP model (G-SGP) to assess the free and occupied spaces and to identify the (\textit{geometric}) navigable space around the robot.
The visual-based and the geometry-based navigable spaces are coupled to calculate a more accurate navigable space, based on which the Local Navigation Points (LNPs) are generated to drive the robot to its destination, see Fig.~\ref{fig_jnt_mdl}.
\begin{figure*}[th!] \vspace{5pt}
\subfloat[Environment\label{fig_sgp_oc_a}]{%
  \includegraphics[width=0.23\textwidth,height=1.in]{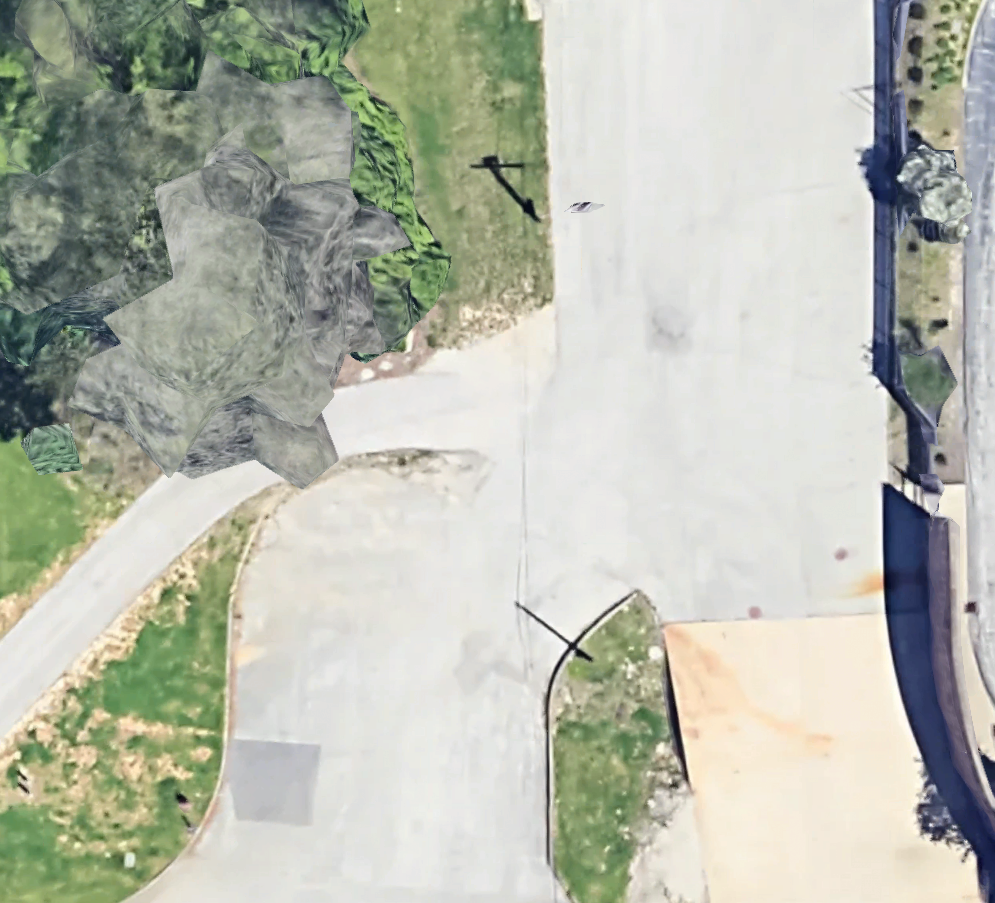}} \hfill
\subfloat[Occupancy surface $\mathcal{S}_{g}$\label{fig_sgp_oc_b}]{%
  \includegraphics[width=0.23\textwidth,height=1.in]{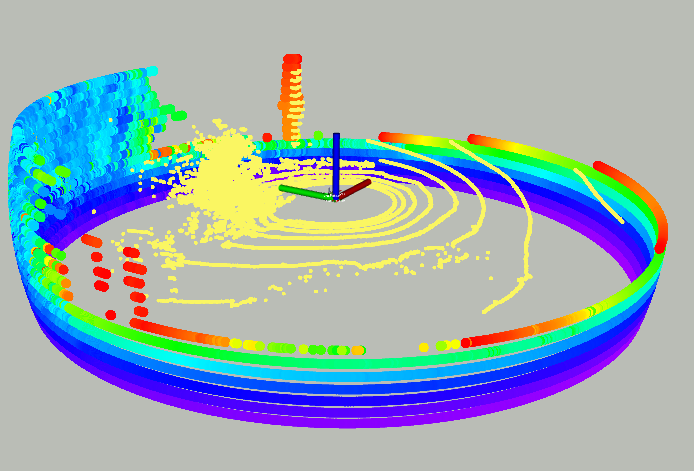}} \hfill
\subfloat[SGP occupancy surface $\mathcal{S}_{\mu_g}$\label{fig_sgp_oc_c}]{%
  \includegraphics[width=0.23\textwidth,height=1.in]{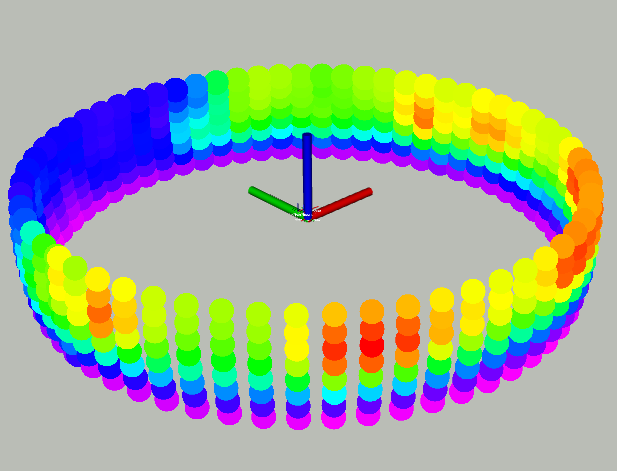}}
  \includegraphics[width=0.01\textwidth,height=1.in]{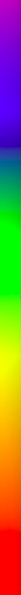} \hfill
\subfloat[Variance Surface $\mathcal{S}_{\sigma_g}$\label{fig_sgp_oc_d}]{%
  \includegraphics[width=0.23\textwidth,height=1.in]{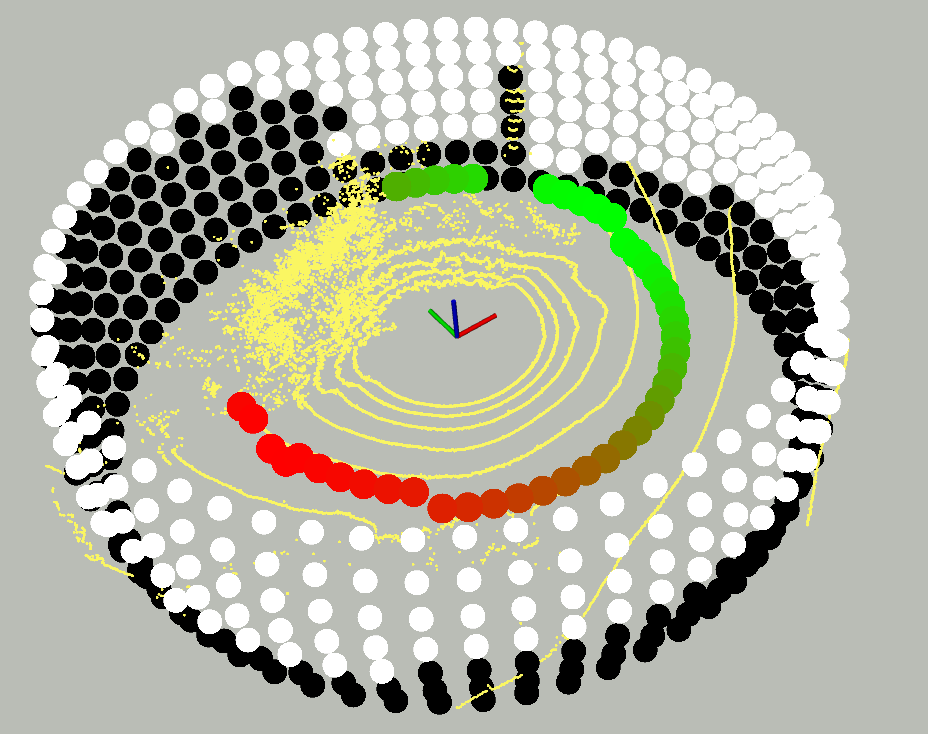}}
  \includegraphics[width=0.01\textwidth,height=1.in]{mthd/red-green-2.png} \hfill
\vspace{-5pt}
  \caption{\small  Geometry-Navigable Space: (b) shows the raw pointcloud in yellow and the original occupancy surface 
  , where warmer colors indicate less occupancy; (c) shows the predicted occupancy surface using the G-SGP model; (d) shows the variance surface, i.e., the uncertainty associated with the predicted occupancy, where white color indicates highly uncertain (free) points. The Geometry-LNPs (G-LNPs) are shown as colored circles, where green indicates less go-to-goal cost associated with each G-LNP. 
\vspace{5pt}
  }
  \label{fig_sgp_oc}
\end{figure*}

\subsection{Geometry Navigable Space: The G-SGP Model }\label{sec_sgp_oc_mdl}
The LiDAR's pointcloud, $\mathbf{P}_{l}=\{(x_i, y_i, z_i)\}_{i=1}^{n_l}$, is converted 
to the occupancy surface, $\mathcal{S}_g$, representation~\cite{10161111}, 
where each (geometry) point, $\mathbf{p_g}_i$, on $\mathcal{S}_g$ is defined by its azimuth, $\alpha_i$, and elevation, $\beta_i$, angles, and given an occupancy value, $\Omega_i$, equals to the difference between the surface radius, $\rho_{g}$, and the point radius, $\rho_i$, as follows $\Omega_i = \rho_{g} - \rho_i$. 
The surface regions with projected points represent the {\em occupied space} of  $\mathcal{S}_g$. 
In contrast, other regions with no points represent the {\em free space} of  $\mathcal{S}_g$,
 see Fig.~\ref{fig_sgp_oc_b}.
The $n_g$ projected points on $\mathcal{S}_g$ form the geometric data set $\mathcal{D}_g = \left\{\left(\mathbf{p_g}_{i}, \Omega_{i}\right)\right\}_{i=1}^{n_g}$, where $\mathbf{p_g}_i=(\alpha_i, \beta_i)$, and $\Omega_i$ is the occupancy of $\mathbf{p_g}_i$. 
Subsequently, the geometric data set $\mathcal{D}_g$ is employed to train a 2D variational SGP (G-SGP), to model the probability of occupancy, $f_g(\mathbf{p_g}_i)$, over $\mathcal{S}_g$ as follows: \vspace{-3pt}
\begin{equation} 
    \begin{gathered}
    f_g(\mathbf{p_g}) \sim {SGP_g}\left(m_g(\mathbf{p_g}), k_{g}\left(\mathbf{p_g}, \mathbf{p_g}^{\prime}\right)\right),  \\
    k_{\mathrm{g}}\left(\mathbf{p_g}, \mathbf{p_g}^{\prime}\right)=\sigma_{1}^{2}\left(1+\frac{\left(\mathbf{p_g}-\mathbf{p_g}^{\prime}\right)^{2}}{2 \alpha_{1} \ell_{1}^{2}}\right)^{-\alpha_{1}},
    \end{gathered}
    \label{eq_geo_sgp}    
\end{equation}
where $m_g(\mathbf{p_g})$ is the zero mean function, and $k_{g}\left(\mathbf{p_g}, \mathbf{p_g}^{\prime}\right)$ is a Rational Quadratics (RQ) kernel with a length-scale $\ell_{1}$, a signal variance $\sigma_{1}^{2}$, and a relative weighting factor $\alpha_{1}$. A Gaussian noise $\epsilon_g$ is added to the predicted occupancy to reflect the measurement noise. The probability of occupancy $\Omega_g^*$ for any query point $\mathbf{p_g}^*$ on $\mathcal{S}_g$ is calculated by the GP prediction as follows, \vspace{-3pt} 
\begin{equation}
    \begin{gathered}
    p_g(\Omega_g^* | \bm{\Omega}_g) = \mathcal{N}_g(\Omega_g^* | m_{\bm{\Omega_g}}(\bm{p_g}^*), k_{\bm{\Omega_g}}(\bm{p_g}^*,\bm{p_g}^*) + \sigma_{n_g}^2), \\
    m_{\bm{\Omega_g}}(\bm{p_g})=K_{\bm{p_g}n_g}\left(\sigma_{g_n}^{2} I+K_{n_g n_g}\right)^{-1} \bm{\Omega_g}, \\
    k_{\bm{\Omega_g}}\left(\bm{p_g}, \mathbf{p_g}^{\prime}\right)=k\left(\bm{p_g}, \bm{p_g}^{\prime}\right)-K_{\mathbf{p_g} n_g}\left(\sigma_{n_g}^{2} I+K_{n_g n_g}\right)^{-1} K_{n_g \bm{p_g}^{\prime}},
    \end{gathered}
    \label{eq_posterior_mean_kernel_full_gp}
\end{equation}
where $m_{\bm{\Omega_g}}(\bm{p_g})$ and $ k_{\bm{\Omega_g}}\left(\bm{p_g}, \bm{p_g}^{\prime}\right)$ are the posterior mean and covariance functions~\cite{titsias2009variational}, 
$K_{n_gn_g}$ is $n_g \times n_g$ co-variance matrix of the inputs, $K_{p_gn_g}$ is $n_g$-dimensional row vector of kernel function values between $\bm{p_g}$ and the inputs, and $K_{n_gp_g} = K_{p_gn_g}^T$. 
We leverage the variational SGP approach~\cite{titsias2009variational} to estimate the kernel hyperparameters $\Theta$ and to select the inducing points $X_m$, more details about the implementation of the G-SGP model can be found in our previous work~\cite{10161111}.
Fig.~\ref{fig_sgp_oc_c} shows the predicted occupancy $\mu_g$ on predicted occupancy surface $\mathcal{S}_{\mu_g}$, where the prediction uncertainty $\sigma_g$ is shown as the variance surface $\mathcal{S}_{\sigma_g}$ in Fig.~\ref{fig_sgp_oc_d}. 
Regarding the accuracy of the SGP occupancy model, the reconstructed pointcloud from a G-SGP model with 400 inducing points has an average error of \SI{12}{\centi\metre}~\cite{10161111}.

 The variance surface $\mathcal{S}_{\sigma_g}$ discriminates efficiently between the free space (white regions with a variance higher than a threshold $V_{g_{th}}$) and the occupied space (dark regions with a variance less than $V_{g_{th}}$) around the robot~\cite{10161230, ali2023autonomous}, and reflects the terrain elevation (the boundary between free and occupied space) in the local observation~\cite{jardali2024autonomous}. Therefore, $\mathcal{S}_{\sigma_g}$ is used to define a set of geometrical-feasible LNPs (G-LNPs) around the robot in the free space that are considered navigable based on the robot's maximum roll and pitch angles. 
G-LNPs are the lowest free points on the variance surface whose elevation angles are bounded by the safe elevations that the robot can climb, ($\beta_{min_s}$ and $\beta_{max_s}$);
$\text{G-LNPs}={(\alpha_i,\beta_{j^*})} \, | \,  -\pi< \alpha_i< \pi  \}$; where $ \beta_{min_s} < \beta_j^* < \beta_{max_s}$.

Formally, a G-LNP is defined as $\mathbf{g}_{lnp_i}=(\alpha_i, \beta_i, \rho_i)$, where  $\alpha_i$ defines the direction of $\mathbf{g}_{lnp_i}$ with respect to the robot heading, $\beta_i$ is the elevation of $\mathbf{g}_{lnp_i}$ with respect to the robot, and $\rho_i$ is the distance between $\mathbf{g}_{lnp_i}$ and the robot predicted by the G-SGP model; $\hat{\rho_i}=\rho_g - \hat{\Omega_i}$, where $\hat{\Omega_i}=\mu_{g_i}$ and $(\mu_{g_i}, \sigma_{g_i})= \mathcal{SGP}_g((\alpha_i, \beta_i))$. 
The Cartesian coordinates of $\mathbf{g}_{lnp_i}$ within the global world frame $\mathcal{W}$ are derived as $(x^{w}_i, y^{w}_i, z^{w}i) = \prescript{W}{}{\mathbf{T}}{R} \cdot (x^{R}_i, y^{R}_i, z^{R}i)$, where $\prescript{W}{}{\mathbf{T}}{R}$ represents 
the robot's localization. The coordinates $(x^{R}_i, y^{R}_i, z^{R}i)$ denote the position of $\mathbf{g}{lnp_i}$ in the robot frame $\mathcal{R}$, calculated from its spherical coordinates $(\alpha_i, \beta_i, \rho_i)$.
\begin{figure*}[th!] \vspace{-5pt}
{
\subfloat[Colored Pointcloud\label{fig_vis_mdl_a}]{%
\includegraphics[width=0.24\textwidth,height=0.9in]{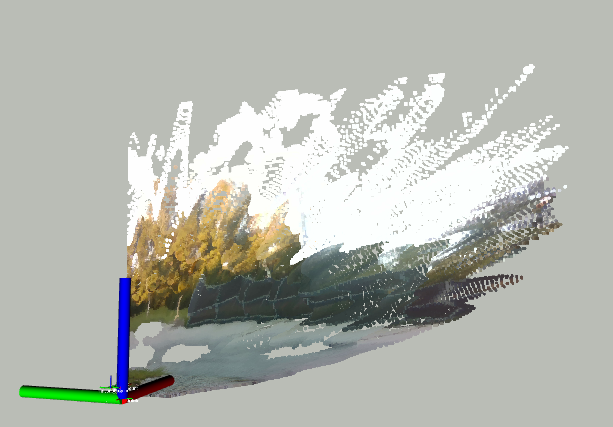}} \hfill
\subfloat[RGB Image\label{fig_vis_mdl_b}]{%
\includegraphics[width=0.24\textwidth,height=0.9in]{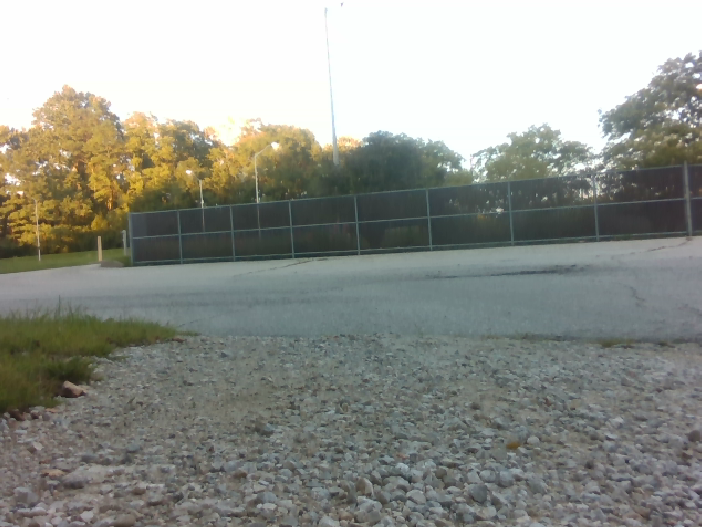}} \hfill
\subfloat[Segmented Image\label{fig_vis_mdl_c}]{%
\includegraphics[width=0.24\textwidth,height=0.9in]{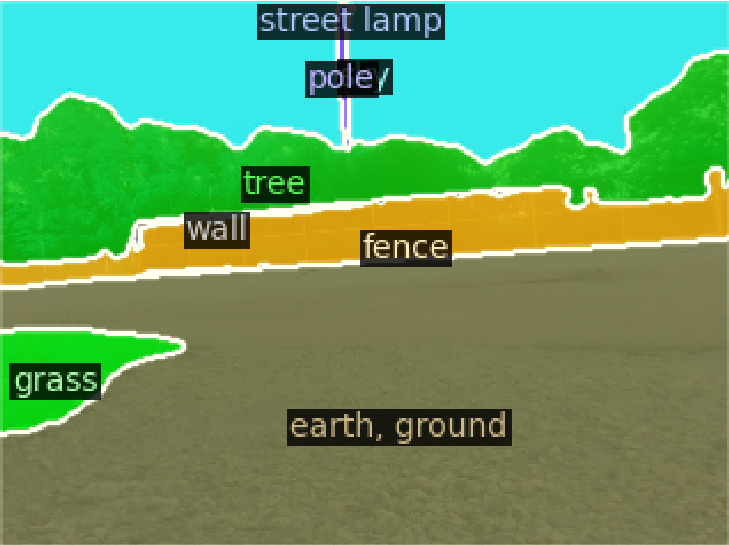}} \hfill
\subfloat[Navigability Image\label{fig_vis_mdl_d}]{%
\includegraphics[width=0.24\textwidth,height=0.9in]{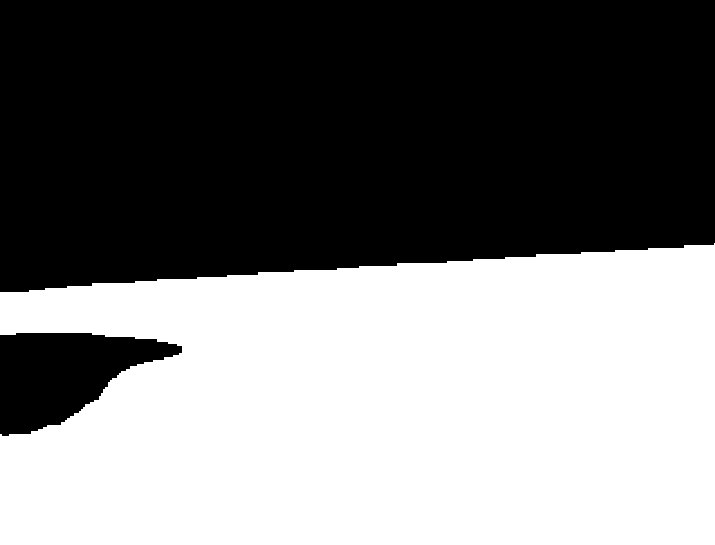}} \hfill  \vspace{-5pt}
  \subfloat[Visual  Pointcloud\label{fig_vis_mdl_e}]{%
\includegraphics[width=0.24\textwidth,height=0.9in]{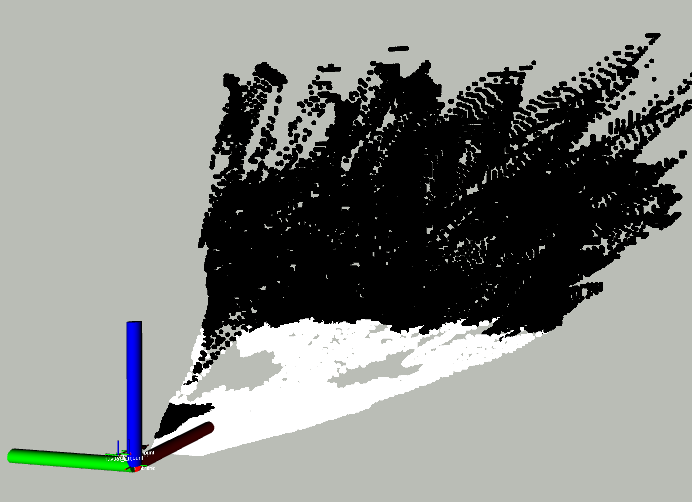}} \hfill
\subfloat[Visual Surface\label{fig_vis_mdl_f}]{%
\includegraphics[width=0.24\textwidth,height=0.9in]{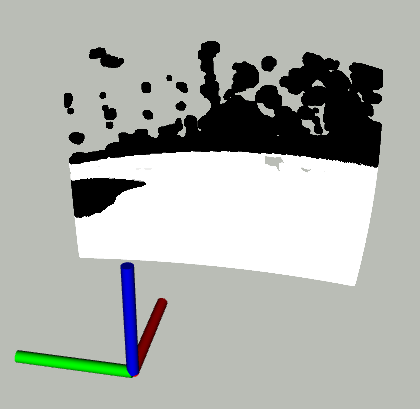}} \hfill
\subfloat[SGP Navigability Model\label{fig_vis_mdl_g}]{%
\includegraphics[width=0.24\textwidth,height=0.9in]{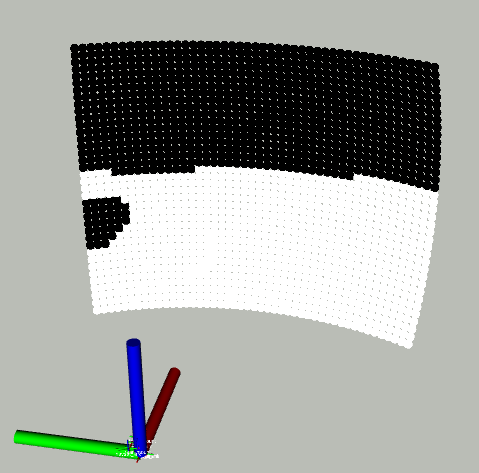}} \hfill
\subfloat[Visual LNPs\label{fig_vis_mdl_h}]{%
\includegraphics[width=0.23\textwidth,height=0.9in]{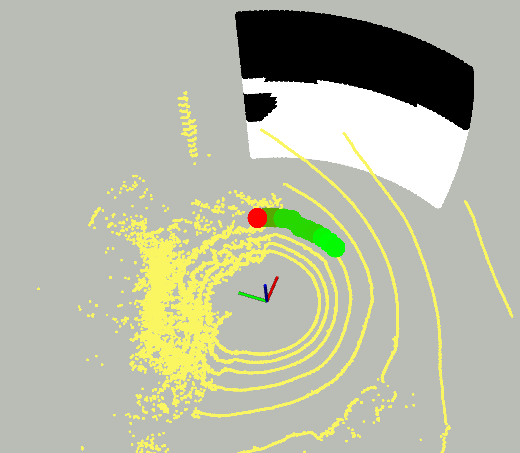}} 
\includegraphics[width=0.01\textwidth,height=1.0in]{mthd/red-green-2.png} \vspace{-5pt}
  \caption{\small  Visual-Navigable Space: V-LNPs are shown as colored-circles in (h), where the color shows the cost assigned to each V-LNP.  \vspace{-5pt}
  }
  \label{fig_vis_mdl} 
  }
\end{figure*}
\vspace{-3pt}
\subsection{Visual Navigable Space: V-SGP Model}
The RGB image contains crucial information contributing to obstacle identification, which may not be revealed through LiDAR sensing (geometry-navigable space). 
The semantic image $I_{seg}$ and the associated class labels $I_{cls}$ can be obtained by $I_{seg}, I_{cls} = g(I_{rgb}, \Theta)$ where $g(.)$ represents 
any existing image segmentation model with parameter $\Theta$, see Fig.~\ref{fig_vis_mdl_c}. In this paper we use the {\em mask2former} segmentation model\cite{cheng2022masked}. 
Utilizing our domain knowledge regarding the physical properties of objects within the scene, we construct the navigability image, denoted as $I_{nav}$, by assigning a value  $0$ to all pixels whose class labels are categorized as non-navigable and a value $255$ to all other pixels, see Fig.~\ref{fig_vis_mdl_d}.
The navigability image provides a dynamic categorization of navigable classes which is crucial in defining flexible visually navigable spaces. Depending on the robot's capabilities and the required behavior, various terrains such as grass, snow, or dirt can be designated as navigable or non-navigable, enhancing the robot's adaptive behavior across different scenarios.

\begin{figure} \vspace{5pt}
{
    \centering  
  \subfloat[D-SGP\label{fig_vis_dpth_mdl_a}]{
      \includegraphics[width=0.46\linewidth,height=1.0in]{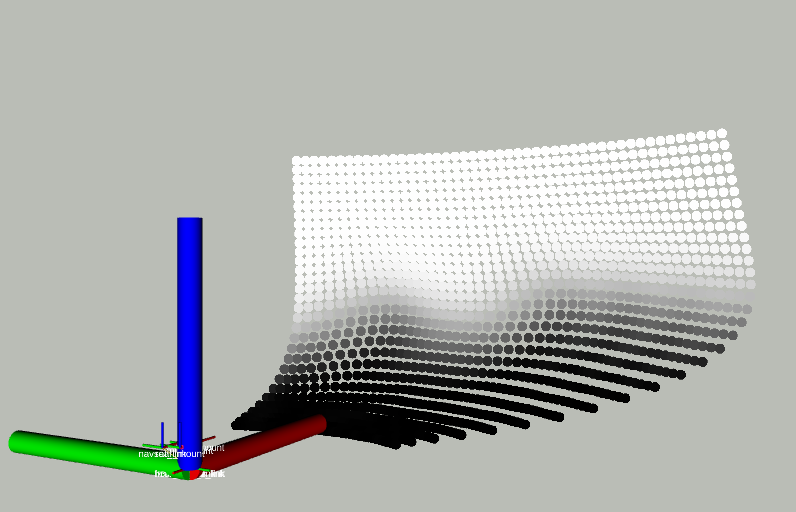} } \hfill
 \subfloat[Combined V-SGP and N-SGP\label{fig_vis_dpth_mdl_b}]{
      \includegraphics[width=0.47\linewidth,height=1.0in]{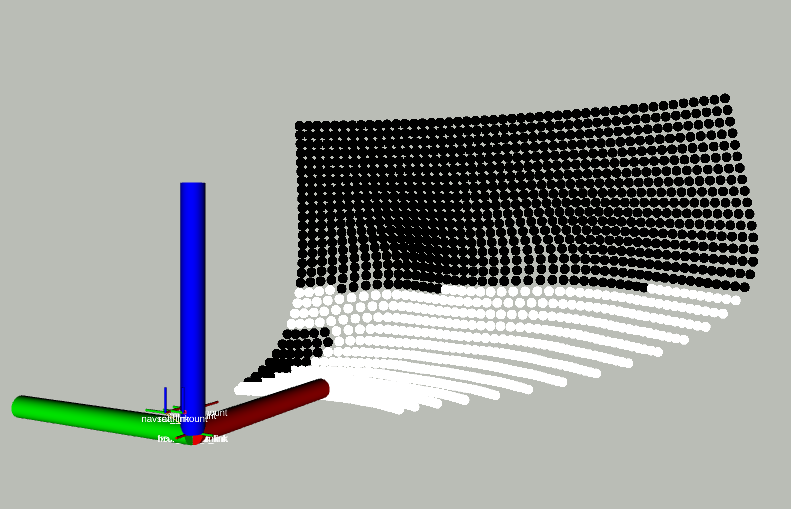} }  \vspace{-5pt}
  \caption{\small D-SGP and D-SGP models of camera's pointcloud: both (a) and (b) shows the pointcloud generated from the D-SGP model, wherein the grey color indicates the uncertainty in (a) and the predicted navigability class in (b). \vspace{-20pt}}
  \label{fig_vis_dpth_mdl} 
}
\end{figure}

The navigability image ${I}_{nav}$ is projected on the visual depth pointcloud $\mathbf{P}_{c}=\{(x_i, y_i, z_i, rgb_i)\}_{i=1}^{n_v}$ by replacing the $rgb_i$ value with the binary navigability value, $\iota_i$, where each point is set to either navigable or non-navigable, resulting in the \textit{navigability pointcloud} $\mathbf{P}_{v}=\{(x_i, y_i, z_i, \iota_i)\}_{i=1}^{n_v}$, see Fig.~\ref{fig_vis_mdl_e}.
Similar to $\mathcal{S}_g$ in Sec.~\ref{sec_sgp_oc_mdl}, the navigability points are projected on a curved visual surface $\mathcal{S}_{v}$ with a predefined radius $\rho_v$, where the visual surface span is aligned with the camera field of view (FoV).
Each point, $\mathbf{p_v}_i$, on $\mathcal{S}_{v}=\{(\alpha_i, \beta_i, \rho_i, \iota_i)\}_{i=1}^{n_v}$, is represented by its azimuth, elevation, radius, and navigability values, see Fig.~\ref{fig_vis_mdl_e}. 
$\mathcal{S}_{v}$ is then decomposed into two data sets, the visual navigability dataset  $\mathcal{D}_{nav}=\{\mathbf{p_v}_i, {\iota}_i\}_{i=1}^{n_v}$ 
and the visual depth data set  $\mathcal{D}_{dpth}=\{\mathbf{p_d}_i, {\rho}_i\}_{i=1}^{n_v}$, where $\mathbf{p_d}_i = (\alpha_i, \beta_i)$. $\mathcal{D}_{dpth}$ is then filtered to include only the navigability points whose $\rho$ less than a constant $\rho_d$, i.e., 
$\mathcal{D}_{dpth}=\{\mathbf{p_d}_i, {\rho}_i\}_{i=1}^{n_v} :  \mathbf{p_d}_i = \mathbf{p_v}_i \mid \rho_i < \rho_d$. 
Thereafter, the visual surface is modeled using two SGP models. The first one is the SGP Depth (D-SGP) model , which is an SGP regression model trained on $\mathcal{D}_{dpth}$ in a similar way as the G-SGP model to predict the occupancy of $\mathbf{p_d}_i$ as  ${\Omega_d}_i = f_d(\mathbf{p_d}_i) + \epsilon_d$: 
\begin{equation}
    \begin{gathered}
    f_d(\mathbf{p_d}) \sim {SGP_d}\left(m_d(\mathbf{p_d}), k_{d}\left(\mathbf{p_d}, \mathbf{p_d}^{\prime}\right)\right),  \\
    k_{\mathrm{d}}\left(\mathbf{p_d}, \mathbf{p_d}^{\prime}\right)=\sigma_{2}^{2}\left(1+\frac{\left(\mathbf{p_d}-\mathbf{p_d}^{\prime}\right)^{2}}{2 \alpha_{2} \ell_{2}^{2}}\right)^{-\alpha_{2}},
    \end{gathered}
    \label{eq_dpth_sgp}    
\end{equation}
where $m_d(\mathbf{p_d})$, $k_{d}\left(\mathbf{p_d}, \mathbf{p_d}^{\prime}\right)$, and $\epsilon_d$ are similar to Eq.~\eqref{eq_geo_sgp}
On the other hand, the second model is the SGP Navigability (N-SGP) model, which is an SGP classification model trained on $\mathcal{D}_{nav}$ to predict the navigability status $\iota$ of any point $\mathbf{p_v}_i$.
The N-SGP classification model uses the same variational SGP framework as the G-SGP model where ${\Omega_v}_i = f_v(\mathbf{p_v}_i) + \epsilon_v$, 
however, its output probability is thresholded to decide whether $\mathbf{p_v}_i$ is navigable or not, see Fig.~\ref{fig_vis_mdl_g}, 
\begin{equation}
    \begin{gathered}
    f_n(\mathbf{p_v}) \sim {SGP_v}\left(m_v(\mathbf{p_v}), k_{v}\left(\mathbf{p_v}, \mathbf{p_v}^{\prime}\right)\right),  \\
    k_{\mathrm{v}}\left(\mathbf{p_v}, \mathbf{p_v}^{\prime}\right)=\sigma_{3}^{2}\left(1+\frac{\left(\mathbf{p_v}-\mathbf{p_v}^{\prime}\right)^{2}}{2 \alpha_{3} \ell_{3}^{2}}\right)^{-\alpha_{3}},
    \end{gathered}
    \label{eq_dpth_sgp}    
\end{equation}
\begin{equation}
    \begin{gathered}
     \iota_{\mathbf{v}_i} =
        \begin{cases}
            \text{255 "navigable"} & \hat{\Omega_{v_i}} >\Omega_{v_{th}}\\
             \text{0 "non-navigable"} & \text{otherwise.} 
        \end{cases} 
    \end{gathered}
    \label{eq_nav_cls}
\end{equation}
where $m_d(\mathbf{p_d})$, $k_{d}\left(\mathbf{p_d}, \mathbf{p_d}^{\prime}\right)$ are similar to Eq.~\eqref{eq_geo_sgp}, and $ \Omega_{v_{th}}$ is a predefined threshold.
Fig.~\ref{fig_vis_dpth_mdl_a} shows the pointcloud predicted by the D-SGP model, $\hat{\mathbf{P}_{c}}=\{(x_i, y_i, z_i, \sigma_{d_i})\}_{i=1}^{N_c}$, where the grey-color intensity indicates the uncertainty $\sigma_d$ of the predicted radius $\hat{\rho}$, i.e., darker points are more certain. Since the model is trained only on the points whose radius values $\rho_i$ are less than $\rho_d$, the D-SGP prediction in the range of $\rho_d$ is certain. 
That is why the dark (certain) points in Fig.~\ref{fig_vis_dpth_mdl_a} are similar, in terms of 3D location, to the flat region (asphalt and grass) of the camera's point cloud $\mathbf{P}_{c}$ in Fig.~\ref{fig_vis_mdl_a}, while the white (uncertain) points have an inaccurate prediction. 
For more details about the accuracy of the generated pointcloud, check~\cite{10161111}. 
It is worth mentioning that the smoothness property of the GP denoises the pointclouds, where individual points that do not belong to any point cluster are smoothed out. 
Fig.~\ref{fig_vis_dpth_mdl_b} shows $\hat{\mathbf{P}_{c}}$ with the color representing the navigability $\hat{\iota}$ predicted by the N-SGP model; $\hat{\mathbf{P}_{v}}=\{(x_i, y_i, z_i, \iota_i)\}_{i=1}^{N}$.
The prediction combination of D-SGP and N-SGP, $\hat{\mathbf{P}_{v}}$, is a reconstruction of the navigability pointcloud $\mathbf{P}_{v}$ in Fig.~\ref{fig_vis_mdl_e}. Only the certain points whose $\hat{\rho} \leq \rho_d$ and $\sigma_d < V_{d_{th}}$, are considered for planning.

The combination of the D-SGP and the N-SGP models is considered as one entity, called the \textit{visual} SGP (V-SGP) model, see Fig.~\ref{fig_vis_mdl}. The V-SGP model can predict both the 3D location of the missing points in the camera's FoV (points with a `nan' return value) and their navigability status, check the difference
 $\mathbf{P}_{v}$ in Fig.~\ref{fig_vis_mdl_e} and $\hat{\mathbf{P}_{v}}$ predicted by the V-SGP model in Fig.~\ref{fig_vis_dpth_mdl_b}. 
In the geometry-based approach, G-LNPs are selected solely based on the geometry information encoded in $\mathcal{S}_{g}$, 
Contrastingly, $\mathcal{S}_{v}$ introduces a more nuanced layer to the navigable space assessment by incorporating the navigability status denoted as $\iota$.
In the visual-based approach, points falling within the safe elevation boundaries, $\beta_{min_s} < \beta < \beta_{max_s}$, are considered as LNP candidates. 
Then the navigability cost $\iota$ is estimated using the N-SGP model for all candidates to form the 
Visual LNPs (V-LNPs), $\bm{\nu}_{lnp}$, see Fig.~\ref{fig_vis_mdl_h}. 
\vspace{-5pt}
\subsection{LNPs Selection For Goal-oriented Navigation} 
In general, to navigate the robot towards a given goal $\mathbf{g} = (x_f, y_f)$ in $\mathcal{W}$, local mapless navigation approaches use different criteria to select one LNP (i.e., an ideal G-LNP ($\mathbf{g}_{lnp}^*$) or V-LNP ($\mathbf{\nu}_{lnp}^*$)) from the feasible LNPs. 
For example, the FGM~\cite{sezer2012novel} selects the ideal LNP based on the area of the free space around it and its direction to the final goal $\mathbf{g}$. 
While the admissible gap approach~\cite{mujahed2018admissible} selects the closest LNP to $\mathbf{g}$.
We utilize the cost function $C_g\left(\mathbf{g}_{lnp_i}\right)$ introduced in~\cite{jardali2024autonomous}, which accounts for the distance to $\mathbf{g}$, the alignment relative to the robot's heading, and the elevation angle of the LNP,
\vspace{-15pt}
\begin{equation}
 \label{eq_cost_fun} 
    \begin{aligned}
C_g\left(\mathbf{g}_{lnp_i}\right) &=  k_{dst}  d_{tg} + k_{dir} \|\alpha_{i}\|  + k_{elv} \|\beta_{i}\|,  \\
d_{tg} &= \rho_{i} + \sqrt{(x_f-x^w_{i})^2+(y_f-y^w_{i})^2},\\
\mathbf{g}_{lnp}^{*} &=\operatorname{arg}\min_{\mathbf{g}_{lnp_i} \in \text{LNPs}}\left(C_g\left(f_{i}\right)\right), 
\end{aligned}     
\end{equation}
where $k_{dst}$, $k_{dir}, k_{elv}$ are weighting factors. 
The go-to-goal cost $C_g\left(\mathbf{g}_{lnp_i}\right)$ is applied to the G-LNPs, where in the case of the V-LNPs the navigability $\iota$ is used to mask the cost of the visually non-navigable LNPs to the maximum as follow,  
\begin{equation}
    \label{cost2}
    \begin{gathered}
        C_v\left( \bm{\nu}_{lnp_i} \right) =
        \begin{cases}
            C_g\left(\mathbf{g}_{lnp_i}\right) & \text{if } \iota_i = 255,\\
            1 & \text{otherwise.} 
        \end{cases} 
    \end{gathered}
\end{equation}
Once the optimal LNP ($\mathbf{g}_{lnp}^*$ / $\mathbf{\nu}_{lnp}^*$) is selected, a motion command with linear and angular velocities, $v$ and $\omega$, is generated as follows:
$v = k_a \rho_{*} - k_b \|\alpha_{*}\|$, and $\omega = k_c \alpha_{*}$.
$\rho_{*}$ and $\alpha_{*}$ are the attributes of the optimal LNP ($\mathbf{g}_{lnp}^*$ / $\mathbf{\nu}_{lnp}^*$), where
$k_a,\ k_b$ and $k_c$ are tunable coefficients.
\vspace{-5pt}
\subsection{Joint Visual-Geometry Model}
Our integration methodology excludes the D-SGP model from the V-SGP model, thereby establishing two distinct SGP Models: the unaltered G-SGP model and the N-SGP model. This process essentially substitutes the D-SGP model with a segment of the G-SGP model, specifically the area where the FOVs of the camera and LiDAR overlap, see Fig.~\ref{fig_jnt_mdl_b}.
The operational workflow begins with the computation of G-LNPs derived from the G-SGP model. Following this, the N-SGP model is engaged to determine the navigability of the identified G-LNPs to define the Visual G-LNPs (VG-LNPs). 
For goal-oriented navigation task, one VG-LNP is selected based on the same go-to-goal cost function introduced in Eq.~\ref{eq_cost_fun}, however the navigability mask in Eq.~\ref{cost2} is modified in such away that each G-LNP is allocated a navigability score as follows,
\begin{equation}
    \label{cost3}
    \begin{gathered}
        C_{vg}\left( \bm{\nu g}_{lnp_i} \right) =
        \begin{cases}
            (1-k_{nav}) \: C_{g}\left(\mathbf{g}_{lnp_i}\right) & \text{if } \iota_i = 255,\\
            1 & \text{if } \iota_i = 0,\\
            k_{nav} \: C_{g}\left(\mathbf{g}_{lnp_i}\right) & \text{Out of camera's FoV.} 
        \end{cases} 
    \end{gathered}
\end{equation}
where $0.5<k_{nav}<1$ is a parameter allowing the user to control the preference between visually navigable VG-LNPs and G-LNPs outside the camera's FoV. When $k_{nav}=0.5$, there exists no preference between the two, granting them an equal cost of $0.5 \, C_g$. Conversely, setting $k_{nav}>0.5$ augments the propensity to opt for VG-LNPs situated within the camera's FoV.
\section{Experimental Design and Results}
\subsection{Experimental Setup} \label{experimental_setup}
A Clearpath Jackal robot equipped with a VLP-16 velodyne and Realsense D435 camera is used to validate our algorithm in both simulation and real-world scenarios. 
For the simulation experiments, we used the grass-mud environment~\cite{lee2023learning}
with ground-truth localization and an RGB-based segmentation to generate the navigability image from the RGB image.
While for real hardware experiments, we used a {microstrain GNSS/INS} module for localization and the \textit{mask2former}~\cite{cheng2022masked} segmentation.
The Velodyne pointcloud $\mathbf{P}_{l}$ is used to construct the occupancy surface  $\mathbf{S}_{g}$ to train the G-SGP model, where the front Realsense D435 RGBD camera is used to generate the navigability poincloud $\mathbf{P}_{v}$ and construct the visual surface $\mathbf{S}_{v}$ to train the V-SGP model. 
Our algorithm runs in real-time with a frequency of 4 Hz in real-world experiment and 10 Hz in simulation due to the time difference required by RGB-Based and the \textit{mask2former} segmentation. Therefore, we limited the maximum velocity in the realworld experiment to 0.5 m/s.
We assessed our algorithm's effectiveness by comparing it to two established methods. The first baseline is a purely visual-based navigation named \textit{PovNav}~\cite{pushp2023povnav}, that employs $I_{rgb}$ for generating $I_{nav}$, similar to our proposed method. However, it uses image-based visual-servoing for motion command generation. The second baseline is GPFrontiers framework~\cite{10161230, jardali2024autonomous}, local mapless navigation which contrasts PovNav by focusing solely on geometry-based navigation. For clarity, we refer to \textit{PovNav} as V-Nav, GPFrontiers as G-Nav, and our proposed approach as VG-Nav.
\subsection{Simulation Scenarios and Results} \label{sec_sim_results}
\begin{figure}[ht] \vspace{1pt}
\centering
\subfloat{%
    \includegraphics[scale=0.45]{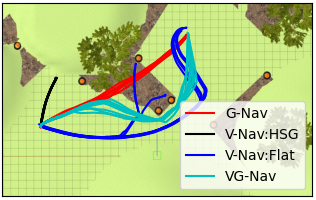}} \hfill
\subfloat{%
    \includegraphics[width=0.19\textwidth,height=1.23in]{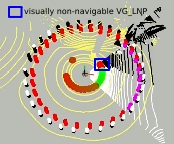}}
    \includegraphics[width=0.01\textwidth,height=1.23in]{mthd/red-green-2.png} \vspace{-15pt}
  \caption{\small  Simulation Experiments: VG-LNPs cost on the right figure. $K_{nav}>0.5$ giving less cost for visually VG-LNPs than G-LNPS outside the Camer'a FoV. 
  }
  \label{fig_sim}  \vspace{-10pt}
\end{figure}

In the grass-mud environment, two visual classes are identified: grass and mud, with grass assigned as navigable and mud as non-navigable. Furthermore, the environment features two types of terrain: flat terrain and high slope grass (HSG), with HSG being geometrically non-traversable.
Our experimental design request the robot to navigate from a start pose of $(15, -4, -\pi/2)$ to a goal pose of $(-4, -16, \pi)$, avoiding mud and HGS regions. Initially facing HSG, the robot must circumvent it, as well as a mud area presented along the straight path to the goal. Successful task completion necessitates the integration of both visual and geometric navigation capabilities to sidestep HSG and mud obstacles. For each approach, we conducted 15 trials to examine its navigation behavior.
First, we executed the G-Nav, which, as anticipated, avoided the HSG area and opted for a direct path to the goal, traversing the mud area due to lack of a visual component. The paths generated by G-Nav are depicted in red in Fig.~\ref{fig_sim}. 
Subsequently, we deployed the V-Nav which failed to circumvent the HSG, perceiving all grass as navigable due to missing the geometric information. 
All 15 V-Nav trials ended with the robot stuck trying to climb the HSG, depicted as black paths in Fig.~\ref{fig_sim}. 
To evaluate the V-Nav performance on flat terrain, we reoriented the robot to face flat areas instead of HSG.
In this revised configuration (V-Nav:Flat), V-Nav successfully navigated through flat grass, avoiding mud areas in 12 out of 15 trials. However, in 3 trials, V-Nav failed to circumvent mud due to erroneous motion commands. 
The critical observation was that V-Nav, when detecting only non-navigable classes (mud) within the camera's FoV, tended to reach a local minimum, failing to avoid going into non-navigable areas, check the blue paths in Fig.~\ref{fig_sim}. 
\begin{table}[h]
\centering
\caption{Navigation metrics. (\textbf{M}: Mud and \textbf{S}: Slope) }
\begin{tabular}{lcccc}
\hline
& Path[m] & $avg(v_{max})[m/s]$ & Avoid \textbf{M} & Avoid \textbf{S} \\
\hline
G-Nav & $22.54 \pm 0.15$ & $1.06 \pm 0.10$ & 0\% & 100\% \\
V-Nav:HSG & - & - & - & 0\% \\
V-Nav:Flat & $33.01 \pm 0.30$ & $0.30 \pm 0.01$ & 80\% & - \\
VG-Nav & $29.21 \pm 0.46$ & $1.03 \pm 0.06$ & 100\% & 100\% \\
\hline
\end{tabular}
\label{table_sim} 
\end{table}
This behavior is replicated in real-world experiments as well, see Fig.~\ref{fig_exp3_a}.
Finally, we tested our proposed VG-Nav in the original setup with the robot facing HSG. VG-Nav successfully guided the robot to the goal while processing both geometric and visual information, effectively avoiding both HSG and mud areas. The resulting paths are illustrated in cyan in Fig.~\ref{fig_sim}.

V-Nav tries to follow the wide visually-navigable space on the navigability image, resulting in a longer path length of $~33$ m, in contrast to VG-Nav, which prioritizes the goal direction unless avoiding mud areas, achieving a shorter average path length of $~29.2$ m. Moreover, the average of the achieved maximum velocity over the 15 trials was $1.06 m/s$ for VG-Nav and $0.3 m/s$ for V-Nav.
VG-Nav, by utilizing a 3D navigability point cloud instead of a 2D navigability image, is more robust in avoiding non-navigable classes, as evidenced in Table~\ref{table_sim}.
\subsection{Real World Scenarios and Results} \label{sec_sim_results}
\begin{figure}[h] \vspace{4pt}
\centering
\subfloat{%
\includegraphics[width=0.22\textwidth,height=0.9in]{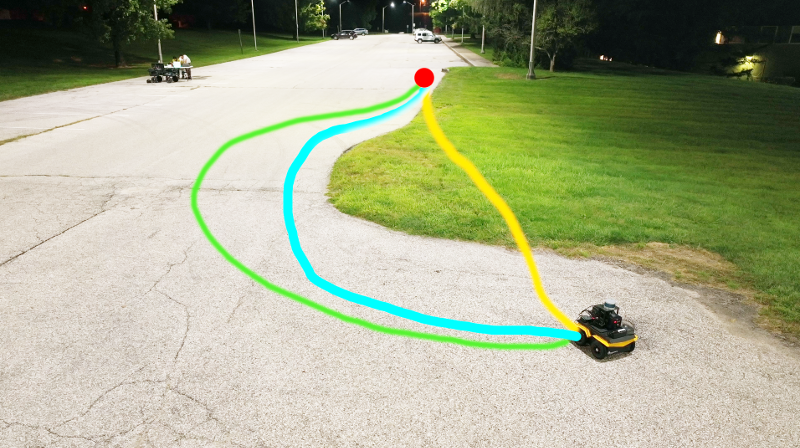}} \hfill
\subfloat{%
\includegraphics[width=0.2\textwidth,height=0.9in]{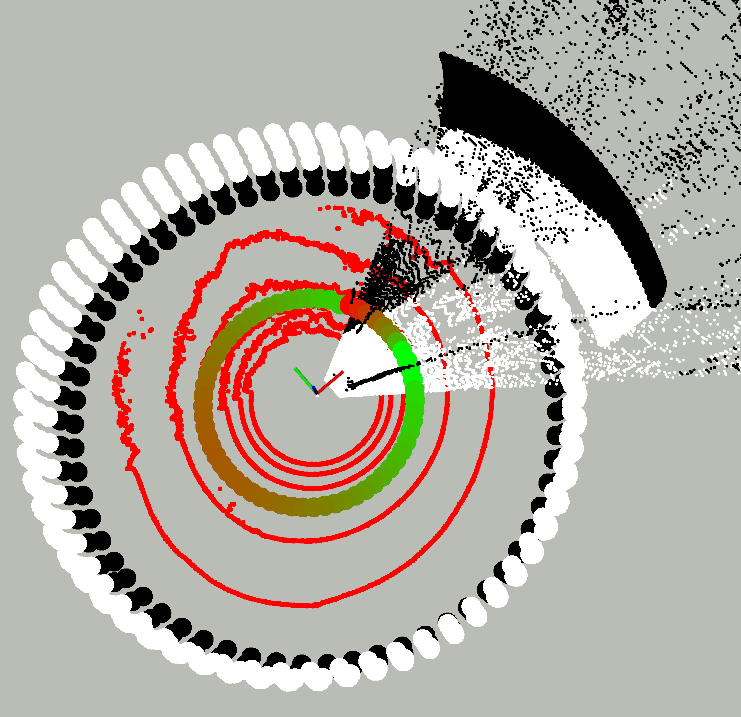}} 
\includegraphics[width=0.01\textwidth,height=0.9in]{mthd/red-green-2.png}  \vspace{-5pt}
  \caption{\small  First real-world experiment: G-Nav can not avoid mud; G-Nav(yellow), V-Nav(green), VG-Nav(cyan). $K_{nav}=0.5$. \vspace{-15pt}
  }
  \label{fig_exp1}
\end{figure}

During real-world experiments, \textit{mask2former}~\cite{cheng2022masked} is used for segmentation. 
While the RGB-based segmentation was robust in simulation for distinguishing mud from grass, \textit{mask2former} shows occasional confusion among similar classes: earth, ground, and grass. This issue is addressed by assigning these four classes identical navigability values.
Notably, \textit{mask2former} reliably distinguished these classes from asphalt/road.
Three experiments were conducted to validate the outcomes of the simulation studies.
In the first setup, the robot is instructed to move towards a goal positioned in a straight line ahead of it. Despite the straightforward path, opting for this route would necessitate crossing over a grassy area, where grass is defined as non-navigable and asphalt/road is defined as navigable.
Subsequently, we conducted 3-trials for each approach. Fig.~\ref{fig_exp1}(left) illustrates the different paths achieved under each algorithm. Notably, the G-Nav failed to avoid the grass, lacking the component to recognize the grass as a non-navigable zone. 
In contrast, both V-Nav and VG-Nav successfully navigated around the grassy region, with VG-Nav opting for a shorter path closer to the grass boundary. These results corroborate the findings from the simulation experiments.
\begin{figure}[h] \vspace{4pt}
{
\subfloat[VG-Nav\label{fig_exp2_a}]{%
\includegraphics[width=0.21\textwidth,height=0.9in]{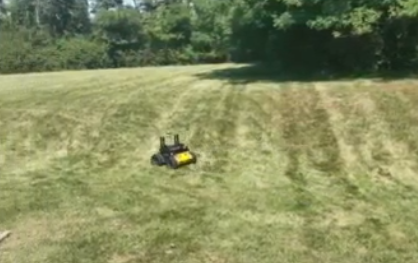}} \hfill
\subfloat[VG-LNPs for (a)\label{fig_exp2_b}]{%
\includegraphics[width=0.21\textwidth,height=0.9in]{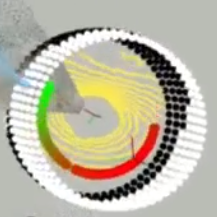}}
\includegraphics[width=0.01\textwidth,height=0.9in]{mthd/red-green-2.png}  \vfill
\subfloat[V-Nav\label{fig_exp2_c}]{%
\includegraphics[width=0.21\textwidth,height=1.0in]{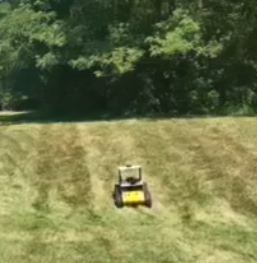}} \hfill
\subfloat[Visual Path for (c)\label{fig_exp2_d}]{%
\includegraphics[width=0.21\textwidth,height=1.0in]{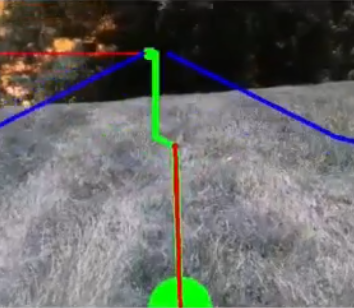}} \vspace{-10pt}
  \caption{\small  Second real-world experiment: V-Nav can not avoid HSG. \vspace{-30pt}
  }
  \label{fig_exp2}}
\end{figure}
\begin{figure}[h] \vspace{4pt} 
{
\subfloat[Paths\label{fig_exp3_a}]{%
\includegraphics[width=0.21\textwidth,height=0.9in]{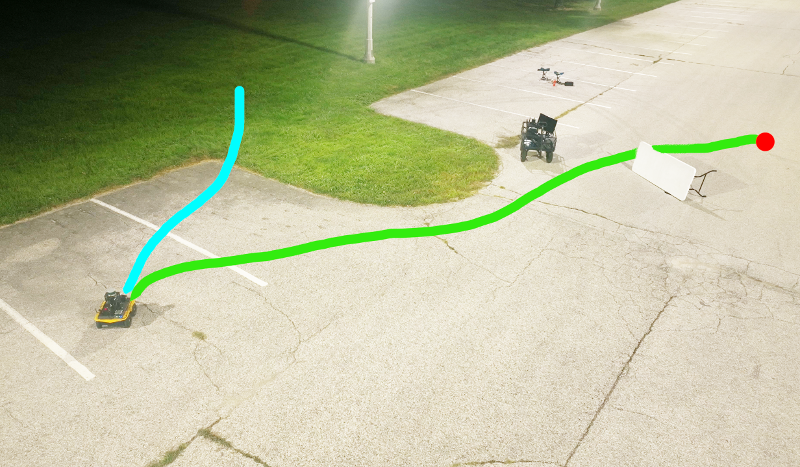}} \hfill
\subfloat[VG-LNPs Cost\label{fig_exp3_b}]{%
\includegraphics[width=0.21\textwidth,height=0.9in]{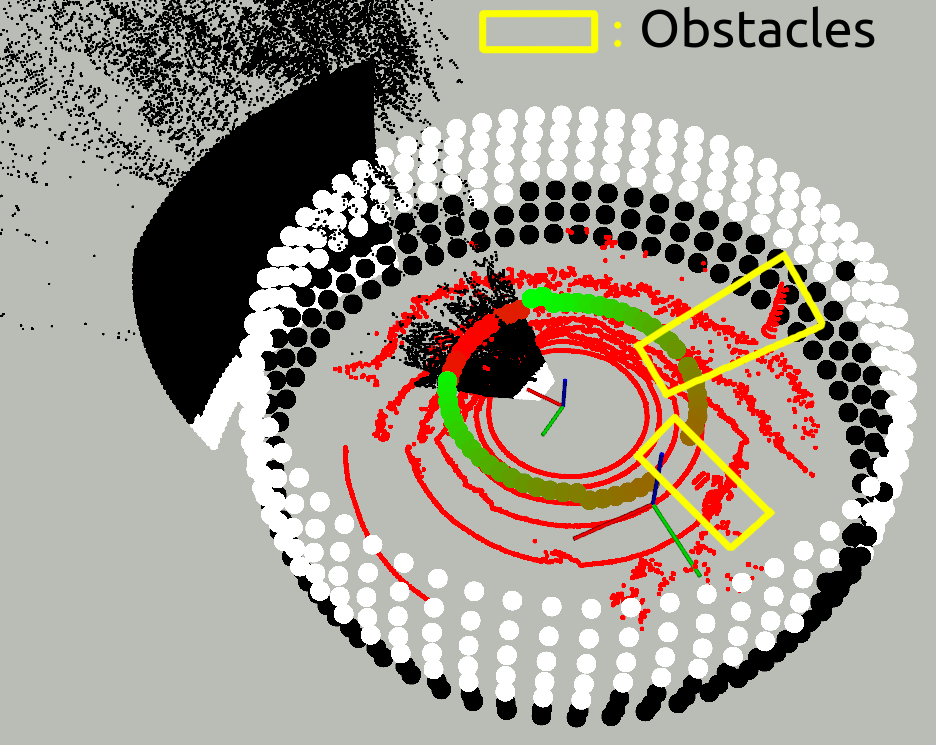}}
\includegraphics[width=0.01\textwidth,height=0.9in]{mthd/red-green-2.png}  \vfill
\subfloat[$v$ and $\omega$\label{fig_exp3_c}]{%
\includegraphics[scale=0.45]{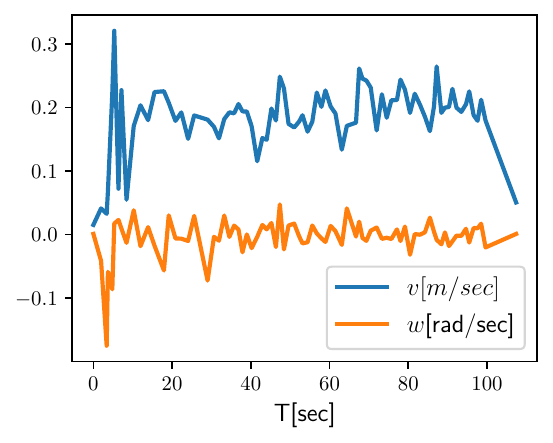}} \hfill
\subfloat[Computation Time\label{fig_exp3_d}]{%
\includegraphics[scale=0.45]{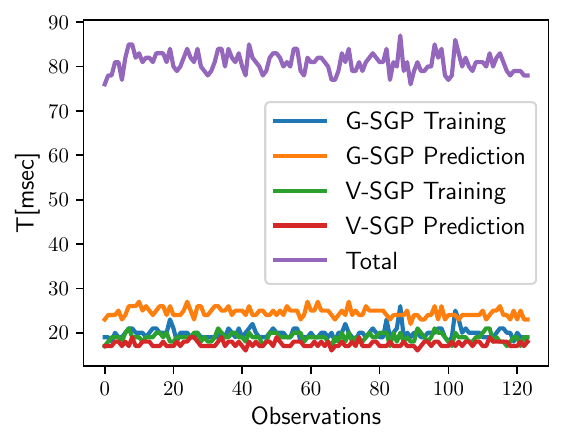}} \vspace{-5pt}
  \caption{\small  Third real-world experiment: V-Nav local minimum due to limited FoV; V-Nav(cyan), VG-Nav(green). \vspace{-20pt}
  }
  \label{fig_exp3}}
\end{figure}
Fig.\ref{fig_exp1}(right) visualizes the cost associated with VG-LNPs. VG-LNPs situated within the non-navigable part of the visual point cloud are assigned the highest cost whereas the remaining VG-LNPs are assigned costs based on the go-to-goal cost function according to Eq.\ref{cost3} with $K_{nav}=0.5$.

In the second experiment, the robot was positioned in an HSG area with grass classified as navigable. The robot was directed towards a goal straight ahead, yet its path passes through an area of grass with the steepest slope. We conducted 3 trials using V-Nav and VG-Nav.
VG-Nav successfully identified HSG and avoided it. As illustrated in Fig.\ref{fig_exp2_b}, which corresponds to the local observation shown in Fig.\ref{fig_exp2_a}, no VG-LNPs are present on the robot's right side that leads to HSG, instead, VG-LNPs are positioned forward, backward, and to the left of the robot, in zones with elevation angles bounded by $\beta_{max}$.
Conversely, V-Nav attempted to guide the robot through the HSG along the straight path. Fig.\ref{fig_exp2_d} depicts the visual path planned based on the local observation in Fig.\ref{fig_exp2_c}. For V-Nav, we ended the experiment before the robot goes to the steepest slope area. These outcomes further validate the insights from the simulation experiments.
The third experiment replicated the local minimum behavior observed with V-Nav.
In this setup, the robot heading was initiated to face a corner formed by a patch of grass (assigned as non-navgiable), deliberately testing the constraints imposed by a limited visual FoV.
Further complicating the path, a table was placed on the direct path to the goal, to introduce a geometric obstacle.
Three trials were conducted, where V-Nav failed to get out of the grass corner in all trials due to the local minimum it encounters when the FoV contains only the non-navigable class, check the cyan path in Fig~\ref{fig_exp3_b}.
In contrast, the VG-Nav demonstrated superior adaptability, managing to identify geometrically feasible G-LNPs outside the camera's FoV. All VG-LNPs in the camera's FoV were masked to the maximum cost, see Fig~\ref{fig_exp3_b}, therefore the VG-Nav command the robot to follow the G-LNP with the minimum cost which lies outside the camera's FOV, this is interpreted from the high angular and low linear velocities values on the first few seconds which led to a sharp rightward rotation, see Fig~\ref{fig_exp3_c}.

We also analyzed the computational costs associated with implementing our VG-Nav by computing the time cost for each individual observation. 
Fig.~\ref{fig_exp3_d} shows the time costs for each component and their sum as the the total time. The average training time of both the G-SGP and the V-SGP models is around $19 msec$, while the average prediction time of the G-SGP ($24 msec$) is higher than 
the average navigability prediction time of the V-SGP model ($17.5 msec$) because V-SGP only predicts the navigability of the feasible G-LNPs identified by the G-SGP model. 
The average total time for each observation is $81 msec$ for the VG-Nav and $43 msec$ for G-Nav, while it is $9 msec$ for V-Nav. 
It is worth mentioning that the time cost for the VG-Nav does not explode over the operation time (number of observations) because we consider each observation individually.
A video highlighting the real-world experiments results is available at: \url{https://youtu.be/0s6VSj5Z1dg}

Contrary to the end-to-end approach, which does not allow the analysis of the individual contributions from geometric and visual information, VG-Nav distinctly employs G-SGP to invalidate all geometrically non-visible G-LNPs and V-SGP to allocate visual traversability costs to the remaining VG-LNPs. Moreover, our methodology is capable of integrating future advanced semantic segmentation techniques to immediately improve navigation behavior without necessitating a training phase.
In the current implementation of VG-Nav, traversability for each terrain type is determined manually, reflecting either the robot's capabilities or preferred navigation behavior, similar to~\cite{ra27, ra33, ra34}. Future developments aim to enhance the V-SGP model's capacity to learn terrain traversability, facilitating adaptation across diverse terrain types in a similar way as~\cite{ra11, ra12, ra13, ra32, ra35}.
\section{Conclusion} \label{conclusion}
We present the Visual Geometry Combined Spaces (VG-SGP) model along with a corresponding navigation strategy (VG-Nav) designed to adeptly guide a robot to its destination. This method leverages the environment analysis of two separate SGP models, G-SGP and V-SGP, to pinpoint areas that are navigable based on both visual and geometric characteristics in the robot's surrounding environment.
By integrating visual and geometric information, our approach facilitates more reliable and adaptive navigation. Simulation and real-world experiments demonstrate that our VG-SGP model and its integrated navigation strategy outperform systems reliant solely on either visual or geometric navigation algorithms, showcasing superior adaptive behavoir required for accomplishing flexible tasks. 


\bibliography{references}

\bibliographystyle{ieeetr}

\end{document}